\documentclass[sigconf]{acmart}
\AtBeginDocument{%
  \providecommand\BibTeX{{%
    \normalfont B\kern-0.5em{\scshape i\kern-0.25em b}\kern-0.8em\TeX}}}

\usepackage{hyperref}
\usepackage{epsfig}
\usepackage{graphicx}
\usepackage{amsmath}
\usepackage[scientific-notation=true]{siunitx}
\usepackage[inline]{enumitem}
\usepackage{tabularx}
\usepackage[ruled,vlined,linesnumbered]{algorithm2e}
\SetKwInOut{Parameter}{Parameters}
\DeclareMathOperator*{\argmax}{argmax}
\newcommand{\eg}{\textit{e}.\textit{g}. }
\usepackage{xcolor}

\copyrightyear{2022}
\acmYear{2022}
\setcopyright{acmcopyright}\acmConference[CODASPY '22]{Proceedings of the Twelveth ACM Conference on Data and Application Security and Privacy}{April 24--27, 2022}{Baltimore, MD, USA}
\acmBooktitle{Proceedings of the Twelveth ACM Conference on Data and Application Security and Privacy (CODASPY '22), April 24--27, 2022, Baltimore, MD, USA}
\acmPrice{15.00}
\acmDOI{10.1145/3508398.3511498}
\acmISBN{978-1-4503-9220-4/22/04}

\begin{document}
\fancyhead{}

\title[Leveraging Disentangled Representations]{Leveraging Disentangled Representations to Improve Vision-Based Keystroke Inference Attacks Under Low Data Constraints}

\author{John Lim}
\email{jlim13@cs.unc.edu}
\affiliation{%
  \institution{UNC Chapel Hill}
  \city{Chapel Hill}
  \state{North Carolina}
  \country{USA}
}

\author{Jan-Michael Frahm}
\email{jmf@cs.unc.edu}
\affiliation{%
  \institution{UNC Chapel Hill}
  \city{Chapel Hill}
  \state{North Carolina}
  \country{USA}
}

\author{Fabian Monrose}
\email{fabian@cs.unc.edu}
\affiliation{%
  \institution{UNC Chapel Hill}
  \city{Chapel Hill}
  \state{North Carolina}
  \country{USA}
}

\renewcommand{\shortauthors}{Lim et al.}

\begin{abstract}
Keystroke inference attacks are a form of side-channel attacks in which an attacker leverages various techniques to recover a user's keystrokes as she inputs information into some display (e.g., while sending a text message or entering her pin).
Typically, these attacks leverage machine learning approaches, but assessing the realism of the threat space  has lagged behind the pace of machine learning advancements, due in-part, to the challenges in curating large real-life datasets.
We aim  to overcome the challenge of having limited number of real data by introducing a video domain adaptation technique that is able to leverage synthetic data through supervised disentangled learning.
Specifically, for a given domain, we decompose the observed data into two factors of variation: \textit{Style} and \textit{Content}.
Doing so provides four learned representations: real-life style, synthetic style, real-life content and synthetic content. 
Then, we combine them into feature representations from all combinations of style-content pairings across domains, and train a model on these combined representations to classify the content (i.e., labels) of a given datapoint in the style of another domain.  
We evaluate our method on real-life data using a variety of metrics to quantify the amount of information an attacker is able to recover. 
We show that our method prevents our model from overfitting to a small real-life training set, indicating that our method is an effective form of data augmentation, thereby making keystroke inference attacks more practical.
\end{abstract}




\maketitle

\section{Introduction}

    We are exceedingly reliant on our mobile devices in our everyday lives. 
    Numerous activities, such as banking, communications, and information retrieval, have gone from having separate channels to collapsing into one: through our mobile phones.
    While this has made many of our lives more convenient, this phenomena further incentivizes attackers seeking to steal information from unsuspecting victims. 
    Therefore, studying  attack vectors and understanding the realistic threats that arise from an adversary's abilities to recover user information is paramount.
    In the case of keystroke inference, the study of such attacks is by no means new; 
    indeed, there is a rich literature of works studying both attacks and defenses. 
    The majority of these attacks utilize  machine learning algorithms to predict the user's keystrokes, \cite{raguram2011ispy}, \cite{ xu2013seeing}, \cite{sun2016visible}, \cite{8418601}, \cite{lim2020revisiting}, but the ability to assess attackers leveraging deep learning methods has lagged behind due to the high costs of curating real-life datasets for this domain, and the lack of publicly available datasets. 
    
    
     \begin{figure*}[!htbp]
        \centering
        \includegraphics[width=\textwidth,height=\textheight,keepaspectratio,]{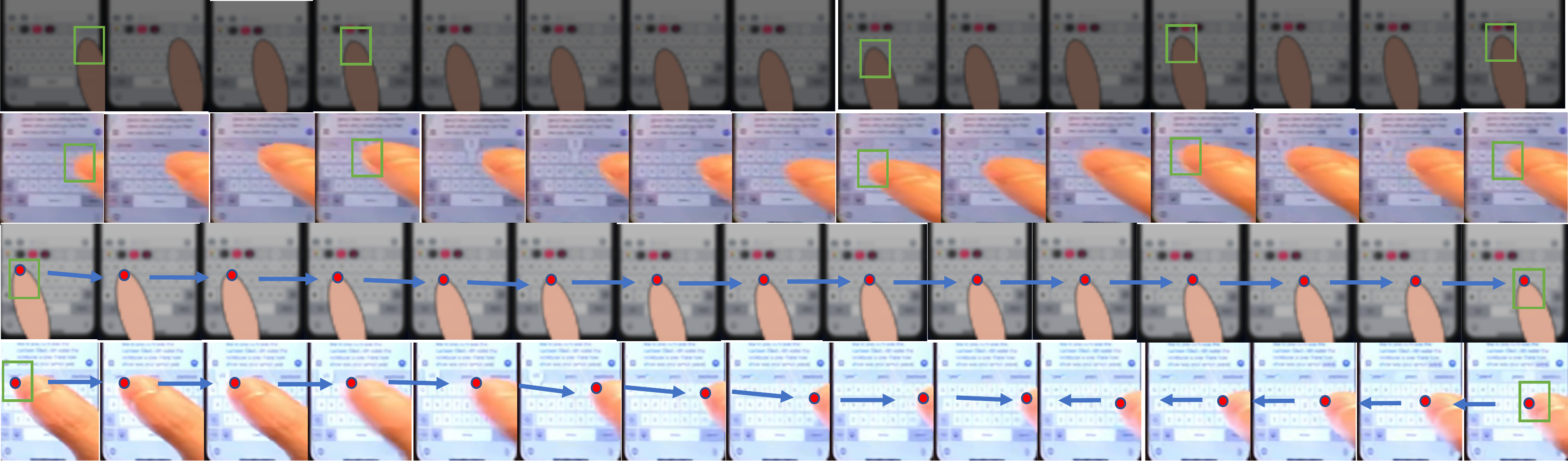}
        \caption{
        An example highlighting the discrepancies between the Synthetic Data (\textit{Rows 1 and 3}) and Real-Life Data (\textit{Rows 2 and 4}). Rows 1 and 2 show sequences of the word \textit{order} being typed with the same number of frames between keypresses sampled.  The frames with  \textcolor{green}{green} boxes indicate ones in which a key was pressed, i.e, in the first frame for first two rows, the key \textit{o} was pressed. While the \textit{content} between the two sequences is the same, the \textit{style} is different, e.g., the texture, and trajectory in between keypresses are different. To further highlight the temporal distribution shift, we show the thumb trajectory between \textit{w} and \textit{h} for both synthetic and real sequences in rows 3 and 4. While the finger is linearly interpolated in the synthetic domain, the real-life one has a more complex one that is challenging to model with a simulator. 
        We highlight the thumb tip in \textcolor{red}{red} and the trajectories in \textcolor{blue}{blue}. 
        }
        \label{fig:Data_examples}

    \end{figure*}

    This paper aims to overcome the challenge of having limited number of labeled, real-life data by introducing a video domain adaptation technique that is able to leverage automatically labeled synthetic data.
    We show that by disentangling our data into separate style and content representations, we can subsequently create style-content pairs across both domains, and combine them into representations that contain the content in the style of its inputs, i.e., style transfer in the feature space. 
    This is especially attractive in the case of pairs of real-life style and synthetic content, as this is an effective data augmentation scheme. 
    Style representations need to be well separated between domains whereas content needs to be indistinguishable. 
    To do this, we introduce auxiliary losses on the latent spaces to enforce disentanglement. 
    Through a series of ablations, we show that doing so improves performance.
    In our context, \textit{Content} answers the question: \textit{What was typed?} 
    (\eg the sentence that a user types).
    \textit{Style} answers the question: \textit{How was it typed?} 
    (\eg the texting pattern).

    Unfortunately, the visual domain adaptation methods available today do not work well in realistic keystroke inference settings because they mainly focus on tasks in which the domain shift is limited to a shift in texture, e.g., image classification, semantic segmentation, etc. \cite{ganin2014unsupervised}, \cite{tzeng2017adversarial}, \cite{hoffman2017cycada}, \cite{motiian2017fewshot}.
    When predicting keystroke sequences, addressing the domain shift with respect to texture is not sufficient. 
    In particular, while there is a clear difference in texture, we must also address the {\it kinematic} domain shift, e.g., different finger motions, speeds, etc.
    Notice, for example, the difference between the trajectories of thumbs in the two example videos displayed in Figure \ref{fig:Data_examples}. 
    The synthetic thumb is linearly interpolated whereas the real one moves in a more complex fashion.

    To summarize, our main contributions are: 
        \begin{enumerate*}[label=\textbf{\arabic*})]
            \item A novel method to assess the threat of keystroke inference attacks by an adversary using a deep learning system while having limited real-life data.
            \item A framework for low-resource video domain adaptation using a supervised disentangled learning strategy that is particularly well-suited to keystroke inference attacks. 
            
        \end{enumerate*}

\section{Background}\label{background}

\paragraph*{Keystroke Inference Attacks}
        Much of the early works in vision-based keystroke inference attacks have focused on direct line of sight and reflective surfaces
        \cite{backes2008compromising, raguram2011ispy, xu2013seeing, yue2014blind,  lim2020revisiting} to infer sensitive data. 
        These data driven approaches are necessary because the attacker can not recover the text using off-the-shelf optical character recognition software (OCR) at low resolutions \cite{yue2014blind}.
        Loosely speaking, the attackers train models that account for various capture angles by aligning the user's mobile phone to a template keyboard. 
        Collectively, these works showed that attackers are able to successfully recover pins and, sometimes, even full sentences. 
        In this paper, we advance the state-of-the-art under the direct line of sight model wherein the attacker uses a mobile camera to record a victim's mobile phone usage. Most germane is the work of 
        Lim et al. \cite{lim2020revisiting} that created a simulator that generates synthetic data for keystroke inference attacks. The authors showed that training with both synthetic and real data, in a supervised domain adaptation framework, yielded a CNN that generalized to a real-life test set. Unfortunately, that work is limited in scope due to the restricted threat scenario they target: analyzing single keypresses.
        By contrast, we assess the ability of an attacker (armed with deep-learning techniques) to recover complete sequences in more demanding settings.

\paragraph*{Style and Content Disentanglement in Videos} 
        Tenenbaum and Freeman (\cite{NIPS1996_1290} \cite{6790155}) observe that by learning to factor observations of data into two independent factors of variation, \textit{style} and \textit{content}, models learn separate representations that can extrapolate style into novel content, classify content in different styles, and translate new content into new styles. 
        Others have disentangled videos into a time-dependent \textit{style} representation and time-independent \textit{content} with adversarial training \cite{denton2017unsupervised}  or with variational autoencoders \cite{li2018disentangled,hsieh2018learning}.  These methods are all unsupervised methods to disentangle style and content. 
        In our setting, style and content are both time-dependent. 
        Style encapsulates {\it the trajectory of the finger in between keys or speed of the user typing. 
        The difference in texture on a per-frame basis is also encapsulated by style. 
        Content represents the entire trajectory as that determines the sentence that was typed. }
        Since we have labels, we take heed of statements made by Locatello et al. \cite{locatello2019challenging}): learning disentangled representations is impossible without supervision, and unsupervised methods using temporal inductive biases do not lead to improved disentangled representations.

 \paragraph*{Low Resource Domain Adaptation}
        We operate in a low resource setting in which we have abundant labels in the source domain and have very few, albeit labeled, data points in the target domain. 
      \citet{hosseiniasl2019augmented} extend the CyCada \cite{hoffman2017cycada} and CycleGAN \cite{CycleGAN2017} frameworks to the low resource domain adaptation setting by adding a semantic consistency loss. 
        \citet{motiian2017fewshot} addresses this problem by learning a feature space that is domain invariant, but is semantically aligned across both domains by introducing a pairing process that augments the datapoints in the target domain.

\paragraph*{Video Domain Adaptation} 
        Domain adaptation for videos has been under explored relative to images, with nearly all methods being limited to human action recognition \cite{Choi2020ShuffleAA}, \cite{pan2019adversarial}. 
        Domain adaptation techniques used for action recognition are not easily transferable to our setting because action recognition methods typically only need a small fraction of frames from the entire video \cite{Schindler2008ActionSH}. 
        However, in our settings, we need to process every frame in order to predict the entire sequence that a user typed. 
        Video translation methods such as \citet{wang2018videotovideo}, \citet{wang2019fewshot}, and \citet{chen2019mocyclegan} show some potential for video domain adaptation tasks, but these methods require that the two domains (RGB images to semantic labels, for example) have the same temporal dynamics.

\section{Our Approach}
    \begin{figure*}[t!]
        \centering
        \includegraphics[width=0.9\textwidth,height=0.9\textheight,keepaspectratio]{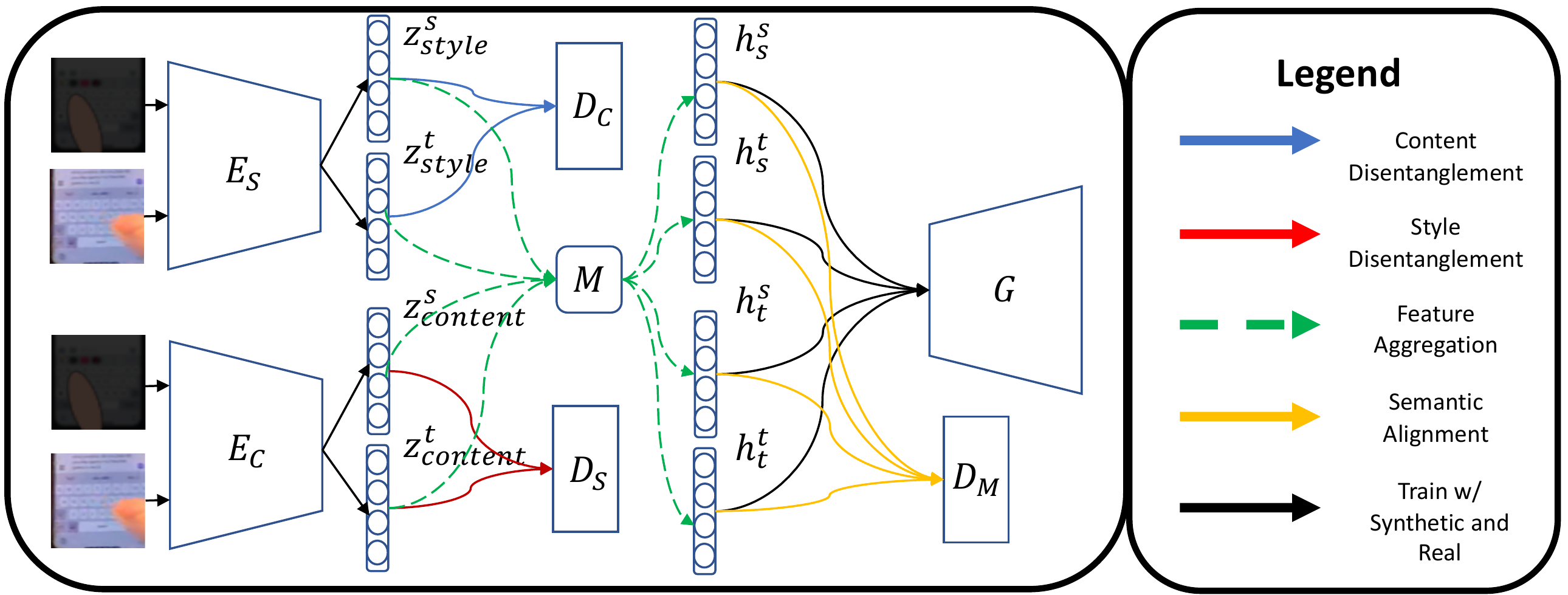}
        \caption{
        A single training iteration takes a pair of synthetic and real-life videos. We disentangle them into style and content representations, and create four combinations of feature representations. For example, real style paired with synthetic content. Style disentanglement, shown in \textcolor{red}{Red}, removes style information from the content space. Content disentanglement, shown in \textcolor{blue}{Blue}, removes content information from the style space. The \textcolor{green}{Green} paths indicate the different ways in which we can combine the style and content representations from the two domains. Finally, we further apply a semantic alignment discriminator to the combined space, shown in \textcolor{yellow} {Yellow}, to ensure the content remains constant, regardless of style. (Best viewed in color)
        }
        
        \label{fig:training_procedure}
    \end{figure*}

     We first provide a brief introduction to keystroke inference attacks and then describe our  framework to disentangle the style and content latent spaces to train on all style-content pairs.
     An overview of our method is shown in Figure \ref{fig:training_procedure}.

\subsection{Keystroke Inference Attacks}
    We model the keystroke inference attack as a Seq2Seq \cite{sutskever2014sequence} problem where the input $X = \{ x_{1}, x_{2}, ..., x_{k} \}$ is a video with $k$ frames and $Y = \{ y_{1}, y_{2}, ..., y_{j} \}$ is a sequence of $j$ characters. 
    The videos are of users typing on their mobile phones that are cropped and aligned to a template image. 
    The tokens are a sequence of characters of the sentence the user typed.
    We do not use any paired data (i.e. the synthetic and real-life datasets do not contain the same sentences), and do not have access to any auxiliary labels such as the exact frame in which a key was pressed.
    Our goal is to learn the parameters of a model that maximizes the conditional probability of $Y$ given $X$.
    We use a Transformer \cite{NIPS2017_7181} encoder-decoder as our model.
    In our setting, we have a dataset of synthetic videos, $\mathcal{D}_s = \{ (X^s_i, Y^s_i) \} $, and a dataset of real-life videos $\mathcal{D}_t = \{ (X^t_i, Y^t_i) \}$, where the number of real-life videos is {\it significantly} less than the synthetic.
    While a large synthetic dataset can be easily generated, there is a distribution shift between the two domains (Figure \ref{fig:Data_examples}).
    Moreover, when the amount of labeled data is scarce, it can be challenging to train neural networks that generalize to samples outside of the training set.

\subsection{Disentangling Style and Content}


     To address the lack of real-life data, we train on combinations of style and content representation pairs from the synthetic and real domains. Additionally,
     we introduce auxiliary losses to enforce disentanglement of style and content, ensuring that the style latent space does not contain any information about the content, and vice versa. 
     Our training framework consists of a Content Encoder, a Style Encoder, a Decoder,  a Feature Aggregation Module, a Style Discriminator,  a Content Discriminator, and a Domain-Class Discriminator (see Fig \ref{fig:training_procedure}). In what follows, we only discuss the intuition and higher level details necessary for understanding how our method works. The loss functions and low-level training specifics are given in the Appendix.

\paragraph*{Pretraining Synthetic Model}

        We first pretrain an Encoder-Decoder Transformer on synthetic data only.
        We train this network with a multi-class cross entropy loss where the goal is to predict the correct sentence for a given video.
        Then the Content Encoder, Style Encoder, and Content Discriminator are initialized with the weights of the pretrained Encoder, and the Decoder is initialized with the weights of the pretrained Decoder.

\paragraph*{Style Disentanglement}

    Style disentanglement ensures that style information is removed from the content latent space.
    The content latent space is defined as the output of the content encoder given a synthetic or real video.
    The content encoder is trained to produce content feature representations that are domain invariant.
    For example, encoding synthetic and real videos of the sentence \textit{``hello, how are you?''} should be close together in the feature space since they have the same semantic information. 
    To achieve this, we train this network in an adversarial fashion \cite{goodfellow2014generative}. 
    Specifically, the Style Discriminator is trained to classify whether a content embedding is real or synthetic, and the Content Encoder is trained to spoof the Style Discriminator.
    Further information can be found in Section \ref{appendix:loss} of the Appendix.

\paragraph*{Content Disentanglement}

        Content disentanglement ensures that content information is removed from the style latent space.
        The style latent space is defined as the output of the Style Encoder given a real or synthetic video. 
        The Content Discriminator is a Transformer Decoder that is trained to predict the correct sentence given the input style representation. 
        The Style Encoder is trained to spoof the Content Discriminator. 
        We do so by producing a style feature representation such that the Content Discriminator can {\it not} predict the correct sentence. 
        We achieve this by maximizing the entropy, $H$, of the predictions of the Content Discriminator.

\paragraph*{Feature Aggregation}
    

        A Feature Aggregation Module combines the disentangled representations from the previous two steps. 
        The aggregation module combines any given pair of style and content embeddings to produce one embedding. 
        For the experiments that follow, we use the LayerNorm \cite{ba2016layer} operation with learnable parameters as our feature aggregation module. 
        There are four different possible pairs that can be the input to our model, since there are two factors of variation (\textit{style} and \textit{content}) and two domains (synthetic and real-life). 
        For any given input pair, the output feature representation can be thought as the \textit{content} in the \textit{style} of the specified domain.

\paragraph*{Prediction}

    The Decoder takes in the output of the feature aggregation module and outputs the predicted sentence, and is trained with cross-entropy loss. 
    At test time, the model outputs the most likely sentence given a real-life video.

\paragraph*{Semantic Alignment}

         Lastly, to further encourage style and content separation, we extend the framework of \citet{motiian2017fewshot} to create training pairs to compensate for limited data in one domain.
         We create four pairs $\mathcal{G}_k, k \in \{1,2,3,4\} $.
        $\mathcal{G}_1$ and $\mathcal{G}_2$ are outputs of $M$ that share synthetic content: (Synthetic Style, Synthetic Content) and (Real Style, Synthetic Content). 
        $\mathcal{G}_3$ and $\mathcal{G}_4$ share real content: (Synthetic Style, Real Content) and (Real Style, Real Content). 
        A multi-class discriminator is trained to correctly identify which group every output of $M$ belongs to. 
        The Content Encoder, Style Encoder, and Feature Aggregation module are trained adversarially such that the multi-class discriminator can not distinguish outputs of the feature aggregation module that are in $\mathcal{G}_1$ and $\mathcal{G}_2$ and outputs of $M$ that are in $\mathcal{G}_3$ and $\mathcal{G}_4$. 
         The high level overview is given in Algorithm \ref{appendix:algo_train}.The revised loss function used to train our model is given in the appendix.

        \begin{algorithm}
            \SetAlgoLined
            \DontPrintSemicolon
    
            \KwIn{Content Encoder, Style Encoder, Feature Aggregation Module, Style Discriminator, Content Discriminator, Decoder, Multi-Class Discriminator, Synthetic Dataset, Real-life Dataset. }

             \While{Not Converged}{
                sample mini-batch of \textit{b} synthetic samples, \{$(X_1^s, Y_1^s)$, $\dotsc$, $(X_b^s, Y_b^s)$ \} from the synthetic dataset. 
                
                sample mini-batch of \textit{b} real-life samples, \{$(X_1^t, Y_1^t)$, $\dotsc$, $(X_b^t, Y_b^t)$ \} from the real-life dataset. 
                
                \textbf{Style Disentanglement}: Remove \textit{Style} information from the Content Space \;

                update the Style Discriminator and Content Encoder \;
                
                \textbf{Content Disentanglement}: Remove \textit{Content} information from the Style Space  \;
                
                update Content Decoder \;
                
                update Style Encoder \;
                
                \textbf{Sequence Prediction} \;
                
                update Content Encoder, Style Encoder, Decoder, and Feature Aggregation Module \; 
                
                \textbf{Semantic Alignment} \;
                
                update Multi Class Discriminator \;
                update Feature Aggregation Module, Content Encoder, and Style Encoder \;

             }
             
             \caption{Learning Algorithm for Disentangling Style and Content. }
             \label{appendix:algo_train}
        \end{algorithm}

\section{Experiments}
    
   Next, we describe the datasets we used, the motivation and interpretation of our chosen evaluation metrics, and our experimental results. To support reproducible research, additional details regarding our data collection methodology and network architectures are given in the Appendix. 
    \begin{figure*}[htp]
        \centering
        \includegraphics[width=1.\textwidth,height=0.5\textheight,keepaspectratio]{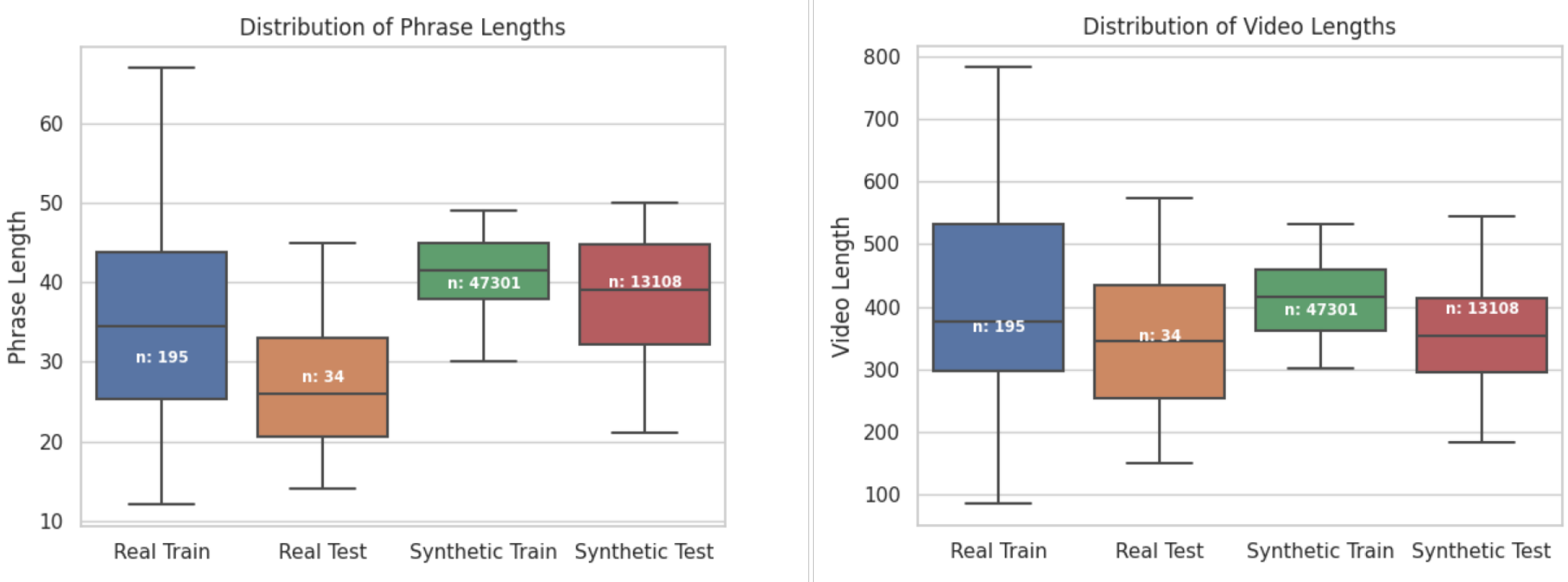}
        \caption{
        Distribution of phrase and video lengths for our datasets. The number of real-life datapoints (229) is significantly less than synthetic (60,409). 
        }
        \label{fig:Dataset_distributions}
    \end{figure*}
    
    \subsection*{Datasets}
        Figure \ref{fig:Dataset_distributions} shows different statistics for the synthetic and real datasets. 
        We set aside 10\% of the training set as a validation set. 
        The real-life dataset was collected by recording participants typing sentences into a mobile phone. 
        Three  participants were asked to type conversational text messages into their mobile devices while we recorded them in both indoor and outdoor settings, with the screen brightness varying according to the environment. 
        We asked the participants to type only with their right thumb.
        We used a mobile camera and captured from a distance of 3 meters. 
        The synthetic data was generated using a simulator \cite{lim2020revisiting} we built. 
        The simulator outputs aligned videos of a synthetic thumb typing a given set of sentences. 
        We generated sentences from the ``A Million News Headlines'' dataset\footnote{https://dataverse.harvard.edu/dataset.xhtml?persistentId=doi:10.7910/DVN/SYBGZL}
        We added a START and STOP token to the beginning and end of a sentence, respectively. 
        In total, there are 30 tokens in which the decoder can predict: 26 letters and 4 special tokens (START, STOP, SPACE, PAD). 
        In both settings, we used a QWERTY keyboard layout. 
    
    \subsection*{Evaluation Metrics}
        We use a variety of metrics to quantify the amount of information the attacker is able to recover from the user because there is no single, agreed-upon, metric for keystroke inference attacks.
        First, we postprocess the outputs of our model with a language model, similar to that done elsewhere \cite{raguram2011ispy, xu2013seeing, sun2016visible, 8418601}. 
        Appropriate metrics for this scenario are Bleu-n \cite{papineni-etal-2002-bleu}, ROUGE \cite{lin-2004-rouge}, and METEOR \cite{banerjee-lavie-2005-meteor}. 
        Bleu-n scores are scored on n-gram precision, i.e., the n-grams in the predicted sentence that are also in the ground truth sentence. 
        ROUGE scores are scored on n-gram recall, i.e., the n-grams in the ground truth that are also in the predicted sentence. 
        METEOR is a metric that is scored on the harmonic mean of unigram precision and recall and was developed to address some of the drawbacks of ROUGE and Bleu-n. 
        METEOR scores range from 0 to 1. 
        Scores above 0.5 reflect understandable translations and scores above 0.7 reflect fluent ones \cite{lavie-2011-evaluating}.
    
        While these scores have merit in the context of keystroke inference attacks, they are not without shortcomings. 
        These scores are especially harsh for predictions that contain slight typographical errors (e.g., ``hello'' vs. ``hellp''), and  there is no guarantee that the previously mentioned postprocessing steps will address every error.
        Also, there are settings in which the applicability of these metrics does not make sense --- e.g., recovering alphanumeric passwords. 
        Thus, we also need evaluation metrics for the raw outputs of our model. 
        Two appropriate metrics are Translation Edit Rate (TER) \cite{Snover06astudy}  and a QWERTY-keyboard-based edit distance. 
        Both metrics measure the number of edits required for a hypothesis sentence to be translated to the ground truth. 
        The latter is a form of the Damerau–Levenshtein (DL) distance \cite{10.1145/363958.363994} that penalizes the edit operations (i,e., insertions, deletions, substitutions, character swapping) conditioned on the QWERTY keyboard layout. 
        For example, if "hello" was the ground truth word, "hellp" should be less penalized than "hellv" as the former is a more likely output than the latter given the assumed keyboard layout.

    \subsection*{Network Architectures} 
        For all experiments, the Encoders ($E_C$, $E_S$) and Decoders ($D_C$, $G$) are both Transformers with 4 layers, 4 attention heads, an embedding size of 128 and a hidden size of 256. 
        $D_M$ and $D_S$ are both 1-layer fully connected layers. 
        Since the output of the Encoder is a sequence of $n$ continuous representations, where $n$ is the input sequence length, we do a max pooling operation along the temporal dimension so that we have a fixed vector representation. 
        These fixed vector representations are the direct inputs to $D_M$ and $D_S$.
        The max sequence length is set at 300, and the max phrase length is set at 70. 
        If an input sequence has more than 300 frames, we randomly sample 300 frames at each epoch. 
        If a video in the testing or validation set has more than 300 frames, we fix the indices of the sampled frames to remove any randomness for evaluation. 
        For input sequences that are shorter than 300 frames, we zero-pad the remaining sequence.


        Table \ref{Baselines_table} shows the results for a model trained and tested on synthetic data. 
        The model performs very well on the synthetic test set across all proposed evaluation metrics.
        To lessen the compute cost of processing over 45k raw videos, we extract a fixed 128-dimensional feature representation as a preprocessing step by training a CNN for single key press classification.
        We use the simulator to generate single key press images and train a CNN to predict the correct key. 

        \begin{table*}[!htbp]
        \centering
        {
            \begin{tabular}{ ccccccc } 
            \hline
            Method & Bleu-1 $\uparrow$ &  Bleu-4  $\uparrow$&  METEOR $\uparrow$ & ROUGE $\uparrow$ & TER $\downarrow$ & Qwerty-D $\downarrow$ \\
            \hline
            
            Synthetic & 0.90 &  0.79 &  .9 & 0.91 & 0.03  &  1.87
            \\ 
             \hline
            Finetuning & 0.15 &  0 &  0.06  & 0.13  & 0.81   &  45.6
            \\ 
            ADDA \cite{tzeng2017adversarial} & 0.15 &  0 &  0.07  & 0.16  &  0.78  & 46.1
            \\ 
            CycleGAN \cite{hoffman2017cycada} & 0.17 &  0 &   0.07 & 0.17  & 0.7  & 45.6
            \\ 
            \hline
      
            Ours (w/o language model) & 0.78 &  0.57 & 0.75  & 0.76  & \textbf{0.09}  &  \textbf{5.3} 
            \\ 
             Ours (full) & \textbf{0.81} & \textbf{0.62}   & \textbf{0.8} & \textbf{0.81}  & \textbf{0.09} & \textbf{5.3}
            \\ 
            \hline
            \end{tabular}
        }
        \caption{\label{Baselines_table} We report various metrics to quantify the attacker's ability to recover information. }
        \vspace{-15pt}

        \end{table*}

\subsection{Baselines}

    To evaluate our approach, we compare against several alternative ideas: finetuning, ADDA, \cite{tzeng2017adversarial}, CyCADA \cite{hoffman2017cycada}, and Vid2Vid \cite{wang2018videotovideo}.
    All methods are evaluated on the real-life test set and use the model trained on synthetic data.
        \begin{itemize}[noitemsep,topsep=0pt,leftmargin=*]
            \item \textbf{Finetuning}. We finetune a model trained only on synthetic with the real-life training set. 
         
            \item \textbf{ADDA}. We use ADDA to generate Encoder output feature representations that are domain invariant to a Discriminator, but are also discriminative for the Decoder. 
            \item \textbf{CyCADA}. 
            We learn a pixel-wise transformation that transforms data from one domain to another.
            We apply this transformation to every real frame to a synthetic one.
            Then, we finetune the synthetic model with the transformed real training set and test on the transformed real test set. 
            Finetuning is needed because \citet{hoffman2017cycada}  do not address the temporal shift as the transformations are conducted on a per-frame basis. 
          
            \item \textbf{Vid2Vid} 
           We leverage a video translation framework that learns to map videos from one domain to another (e.g., labels to RGB images). 
            Our aim is to translate the real-life videos in our training set into synthetic versions, finetune the model trained only on synthetic data, translate the real-life videos in our testing set into real-life versions, and test on the transformed real-life test set.
        \end{itemize}
        
        \subsubsection*{Findings}
        Although we were successful in applying ADDA to the task of single key press classification \cite{lim2020revisiting} when the number of labeled data is scarce,
        we found that simply applying ADDA to our sequence prediction task leads to severe overfitting due to the limited real-life data, indicating that this task is more challenging than the single keypress classification task.
        While CyCADA and ADDA are common approaches for visual domain adaptation, we did not find them suitable for our sequence prediction problem.
        This is because these approaches are tailored to domains in which the domain shift is limited to textures. 
        Recall that in our case, we are facing both a kinematic and texture domain shift.
        This is especially true for CyCADA and other pixel-wise transformation methods.
        We carried out numerous experiments to tune our baselines and maximize their performance, but despite an extensive search of hyperparameters, the models still overfit.
        We report the best results in Table \ref{Baselines_table}. 
        
        Another important fact is that Vid2Vid is trained with pairs that are composed of a video in one domain (RGB) paired with the same with video in another domain (Semantic Labels).  This provides both global and local supervision. 
        By global supervision, we mean that the video for the two domains are of the same event.
        By local supervision, we mean that there is supervision on a per-frame basis. 
        For example, every RGB frame corresponds to a semantic label frame. 
        While such supervision exists for the datasets  (\cite{cordts2016cityscapes, 2020, Rssler2018FaceForensicsAL}) used by Wang et al. \cite{wang2018videotovideo}, we are unable to simulate such supervision.
        We can generate synthetic versions of any real-life video (global supervision), but we are unable to simulate the synthetic thumb trajectories such that the temporal dynamics are the same (local supervision).
        Despite this, we still attempt to generate realistic videos with only global supervision. 
        Our aim is to transform every real-life video into a synthetic one. 
        
        For Vid2Vid, we used the official implementation provided by the authors.\footnote{https://github.com/NVIDIA/vid2vid}
        First, we generate a synthetic video for each of the real-life videos in our training and test sets. 
        Next, we clip the lengths of the videos so that the number of frames is equal for a given (synthetic, real) video pair.
        Then, we train using this set of real and synthetic pairs using the default hyperparameters used in \cite{wang2018videotovideo}. 
        Once the generator is trained, we transform each real-life training video to a synthetic version. 
        We finetune the pretrained synthetic model using the transformed real-life training set. 
        Finally, we test on transformed real-life videos. Even so, this method failed to yield any results as the generator was unable to generate plausible looking synthetic videos -- underscoring the need for a new approach like ours.
    
\subsection{Adapting to Real-Life Videos}

     \begin{table*}[!htbp]
    \centering
    \small
    {\begin{tabular}{ ccccccc } 
    \hline
     Method & Bleu-1 $\uparrow$ &  Bleu-4  $\uparrow$&  METEOR $\uparrow$ & ROUGE $\uparrow$ & TER $\downarrow$ & Qwerty-D $\downarrow$ \\
    \hline
    \textbf{I -} Base & 0.73 &  0.52 & 0.74  & 0.75  & 0.12  &  7.4 \\ 
    \textbf{II -} Base + Style  & 0.77 &  0.56 & 0.76  & 0.77  & 0.1  &  6.3 \\ 
    \textbf{III -} Base + Content & 0.77 &  0.53 & 0.76  & 0.78  & 0.11  &  5.9 \\ 
    \textbf{IV -} Base + Style + Content & 0.76 &  0.57 & 0.75 & 0.76 & 0.12 & 7.0 \\ 
    \textbf{V -}  Base + Semantic Alignment  & 0.77 & 0.57   & 0.76 & 0.79  & 0.11 & 5.7 \\
    \textbf{VI -} Full  & \textbf{0.81} & \textbf{0.62}   & \textbf{0.8} & \textbf{0.81}  & \textbf{0.09} & \textbf{5.3} \\
    \hline
   \textbf{I -} Base (100) & 0.58 & 0.33 & 0.56 & 0.6  & 0.2  &  13.1 \\ 
    \textbf{II -} Base + Style (100) & 0.65 &  0.38 & 0.62 & 0.65 & 0.18 & 11.2 \\ 
    \textbf{III -} Base + Content (100) 
    & 0.62 &  0.37 & 0.57 & 0.61 & 0.2 & 11.3 \\
    \textbf{IV -} Base + Style + Content (100) & \textbf{0.69} & 0.4 & \textbf{0.65} & \textbf{0.69} & \textbf{0.15} & \textbf{9.3} \\ 
    \textbf{V -} Base + Semantic Alignment  (100)  & 0.65 & 0.34   & 0.6 & 0.65  & 0.18 & 11.3 \\
   \textbf{VI -} Full (100) & 0.65 &  \textbf{0.42} &  0.63 & 0.67 & 0.21 & 13.2 \\

    \hline
    
    \end{tabular}
    }
    \caption{\label{Ablation} We conduct ablation studies to evaluate the effectiveness of each loss component. We also evaluate performance when the number of real training videos is dropped to 100 from 175. }
    \end{table*}

        \begin{figure*}[!htbp]
            \centering
            \small
            \includegraphics[width=\textwidth,height=\textheight,keepaspectratio]{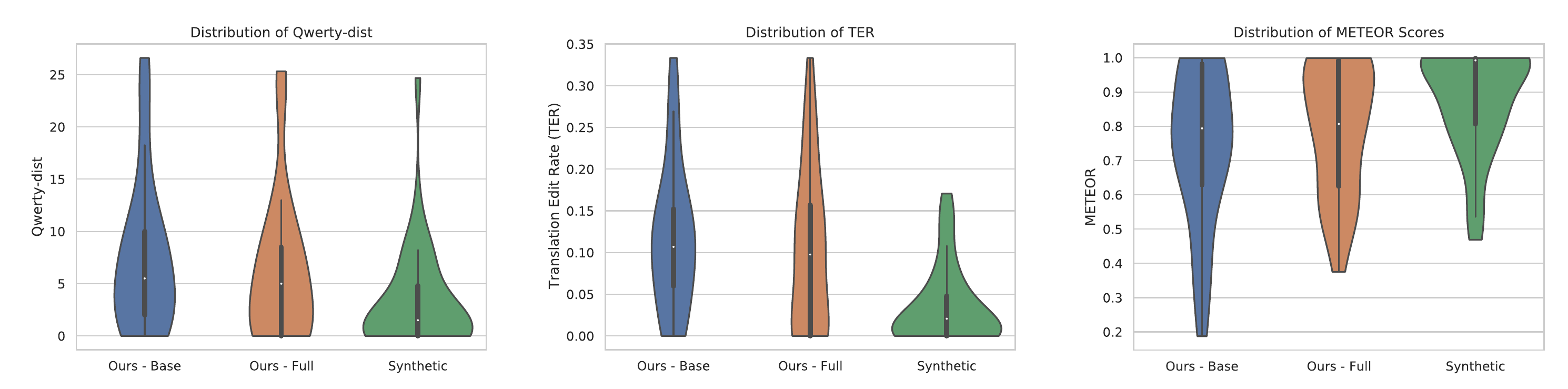}
            \caption{
            From left to right: the distribution of Qwerty-D, TER, and METEOR scores, respectively. ``Full'' is our proposed framework. ``Base'' is our framework without any adversarial training.  We also compare these two against a model trained and tested using only synthetic input.
            }
            \label{fig:violin plots}
        \end{figure*}

Our method, unlike the above baselines, does not overfit to the real-life training set. 
Our results show that training with our pairing mechanism with disentangled representations across domains is an effective form of data augmentation.
We outperform the baselines in both raw output evaluations and post-processed evaluations as shown in Table \ref{Baselines_table}. 
We found that our training was not sensitive to the hyperparameters and weightings of the loss terms (in Equation \ref{eq:full loss} in the Appendix), and use the same hyperparameters for all experiments.

While a direct comparison to the state of the art in direct line of sight attacks is difficult due to the differences in datasets, it is worth noting how our model performs relative to others.
Raguram et al. \cite{raguram2011ispy} achieve a METEOR score of 0.89 whereas Xu et al. \cite{xu2013seeing} achieve a score of 0.71, albeit with recordings taken from much farther distances. 
To measure an attacker's ability to recover passwords, Raguram et al. \cite{raguram2011ispy} report precision and recall for individual word units and characters. 
They achieve word-level precision and recall of 75\% and 78\%, respectively, and character-level scores of 94\% and 98\%.
We achieve a word-level precision and recall of 78\% and 79\%, respectively, and a precision and recall of 96\% and 95\%, respectively, for characters.
\citet{raguram2011ispy} does not report METEOR scores for this scenario.
In their experiments, Raguram et al. \cite{raguram2011ispy} use three different cameras in their experiments: Canon VIXIA HG21 Camcorder, Kodak PlayTouch
and Sanyo VPC-CG20. 
Xu et al. \cite{xu2013seeing}, on the other hand, use a Canon 60D DSLR with 400mm lens and a Canon VIXIA HG21 Camcorder.

\begin{figure*}[!htbp]
    \centering
    \tiny
    \includegraphics[width=\textwidth,height=\textheight,keepaspectratio]{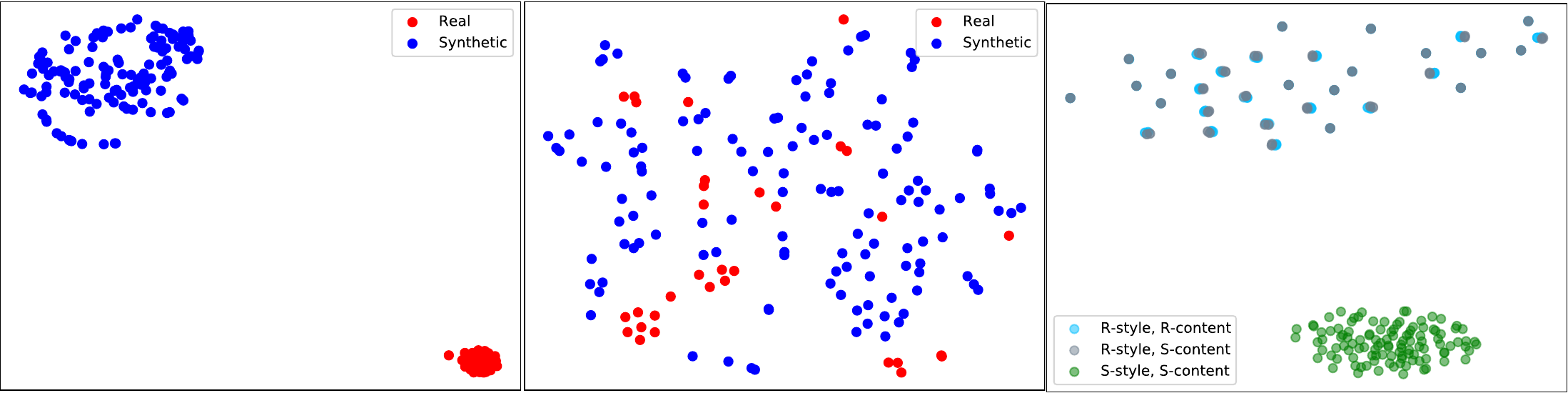}
    \caption{
    t-sne plots for the outputs of $E_S$ (Left), $E_C$ (Center), and $M$ (Right).
    }
    \label{fig:tsne}
\end{figure*}

\subsubsection{Feature Visualization}
For pedagogical purposes, the t-sne  \cite{Maaten2008VisualizingDU} plots in Figure \ref{fig:tsne} show the feature representations of $E_S$, $E_C$, and $M$ on synthetic and real test data. Notice how that sentences with different styles have a noticeable separation, whereas the content representations are intertwined. 
The last figure on the right shows outputs of our feature aggregation module, $M$, and shows the transfer of styles in the feature space. 
Notice the clear separation between styles, while the datapoints within one style cluster are mixed. 
To obtain inputs suitable for t-sne, we perform a max pooling operation along the temporal dimension of the outputs of the networks.

\subsubsection{Ablation}
Lastly, we conducted a series ablation studies to explore the effectiveness of our proposed framework.
We introduce seven different models: \begin{enumerate*}[label={}]
  \item \textbf{I} is our base method without the use of any adversarial losses, just the pairing mechanism.
  \item \textbf{II} uses style disentanglement, i.e., \textbf{I}  $+$ style disentanglement.
  \item  \textbf{III} uses content disentanglement, i.e., \textbf{I} $+$ content disentanglement.
  \item \textbf{IV} uses both style and content disentanglement.
  \item \textbf{V} is the base method with the modified semantic alignment loss.
  \item \textbf{VI} uses style and content disentanglement, along with the semantic alignment. This is our proposed method trained with Algorithm \ref{appendix:algo_train}, i.e., \textbf{IV} $+$ semantic alignment. 
 \end{enumerate*}

First, we find that our base model (\textbf{I}) achieves competitive results without any losses on the latent spaces. 
This indicates that training on paired representations across domains is an effective method for data augmentation.
Second, we find that adding auxiliary losses on the latent spaces to enforce style and content disentanglement improves performance. 
The performance for models \textbf{II} and \textbf{III} shows the base model is benefiting from the added loss terms.
The results for Model \textbf{IV} aligns with our hypothesis that explicitly disentangling style and content allows us to overcome the lack of training data in the target domain by training with all combinations of the factors of variation. 
Finally, we trained model \textbf{V} to apply the semantic alignment step on our paired outputs without any additional adversarial losses. 
This is quite competitive with \textbf{IV}, but we find the greatest performance boost when training model \textbf{VI} using both semantic alignment and disentanglement.
A closer look into the distribution of the scores in Figure \ref{fig:violin plots} shows that the distribution of scores for Model \textbf{VI} (Full) indicates higher overall performance compared to Model \textbf{I} (Base).
{\bf Our results show that explicitly disentangling style and content by adding the adversarial losses on the latent spaces, supplements the pairing mechanism to achieve the highest performance under the evaluation metrics}.

\subsection{Implications}
    Taken as a whole, our results show an adversary seeking to deploy keystroke inference attacks can leverage deep learning methods, despite the difficulties in curating training data. 
    While we are unable to directly show whether such an adversary would outperform those in prior attempts, we show that we can train a deep learning model using the same amount of real-life data used in previous studies. 
Thus, the settings for vision-based keystroke inference attacks should be revisited as the realism and threat capacity of these attacks are most likely greater than initially thought. In particular, the lack of training was a significant impediment to many earlier proposals, but having a synthetic-to-real domain adaptation framework like ours --- to augment the limited real data --- would likely lend itself to a more capable adversary. 
    
    Furthermore, research over the past few years  has shown that deep learning methods have outperformed shallow methods in most computer vision tasks. Arguably, if we hold all the parameters of a threat scenario constant (distance, camera model, phone model), an attacker using a deep learning method should outperform one with a shallow method. Similarly, our style and content disentanglement techniques can be used to help an attacker thwart certain defenses. 
    For example, \citet{sun2016visible} proposed the use of random device perturbations as a way to mitigate an attacker's ability to map backside perturbations to keystrokes. While this is an effective defense, an attacker can undermine it by training a model to disentangle the fake perturbations from the real backside perturbations.

\section{Conclusion}
    Our work provides the important initial step needed to formulate defenses for keystroke inference attacks in the age of deep learning. 
    Specifically, we provide the first assessment of an attacker's ability to recover sentence level information using deep learning systems. 
    We address the problem of limited training data by introducing a framework for low resource video domain adaptation, that disentangles the style and content across both domains, and creates representations from all pairs of style and content combinations. 
    Our results indicate that training with these pairs, along with auxiliary losses to explicitly disentangle style and content, serves as an effective form of data augmentation that prevents overfitting.
    We evaluate our method using a number of metrics to quantify the amount of information the attacker is able to recover, and our results show that an attacker armed with a deep learning system is able to recover enough information to pose a significant threat to unsuspecting victims.
    Our framework can also be used to assess other keystroke inference attacks such as those that focus on device perturbations \citep{sun2016visible} or eye gaze \citep{8418601}. 
     
\section{Availability}
    The code and data used in this paper can be found at \url{https://github.com/jlim13/keystroke-inference-attack-deep-learning}.



\bibliographystyle{ACM-Reference-Format}
\bibliography{sample-base}

\appendix

\section{Additional Training Details}\label{appendix:additiona details}





\subsection{Synthetic Single Key Press Classifier}\label{appendix:syn_single_key}

    We train a CNN, $\phi(\cdot)$, for the task of single key press classification in order to learn a $d-$dimensional ($d = 128$) feature extractor. 
    Once this network is fully trained for the task of single key press classification, we can extract the features of each video on a per-frame basis. We use the simulator by \citet{lim2020revisiting} to generate 70,000 single key press images. 
        These images contain the synthetic thumb over one of 27 keys on the QWERTY keyboard (26 letters $+$ the space bar). 
        Once generated, these images are preprocessed in a similar fashion to the synthetic video dataset. 
        We resize the images to size 200 X 100 and crop the phone such that only the keyboard is showing. 
        We use 50,000 images for training and 10,000 images for testing and validation, respectively. We use a CNN where each layer consists of a Convolution Layer, ReLU activation, and MaxPool operation. 
        We use 3 layers and 2 fully connected layers. 
        We train a network to achieve 95\% accuracy on a held out test set without much hyperparameter or architecture search, as this is a fairly simple 27-way classification task. 
        We use this final model to preprocess every frame in our synthetic video dataset. 
        Every video is a now a sequence of these $d-$dimensional feature representations.
        We use the Adam optimizer with a learning rate of 0.0001.

\subsection{Real-Life Single Key Press Classifier}\label{appendix:real_single_key}

    When extracting the visual features for the real-life videos, we can not use a feature extractor that was trained only on synthetic data. 
    There is a distribution shift between the synthetic and real-life data, so the features we extract would be not be informative.
    Instead of using $\phi(\cdot)$ that was trained for single keypress classification on just synthetic data, we train $\phi(\cdot)$ with a combination of synthetic and real-life data.
    Specifically, we adopt the ADDA \cite{tzeng2017adversarial} framework for unsupervised domain adaptation to train $\phi(\cdot)$. 
    We treat the individual frames for all of the videos in our real-life training set as unlabeled data.
        Even though we do not have labels for individual keypresses for real-life data, we can leverage the fact that we have abundant labels for synthetic data by adopting the unsupervised domain adaptation technique ADDA.
 We use the CNN for single key press classification on synthetic data as our pretrained network. 
        The Discriminator is a 1 layer, 128-dimensional fully connected layer followed by a sigmoid.
        We follow the same guidelines to train ADDA as the original paper \cite{tzeng2017adversarial}, and refer the reader to this work for the full description of their training process.
        We use the Adam optimizer and a learning rate of 0.0001 for both the Discriminator and CNN.

\subsection{Network Architectures}\label{appendix:network architectures}

    For all experiments, the Encoders ($E_C$, $E_S$) and Decoders ($D_C$, $G$) are both Transformers with 4 layers, 4 attention heads, an embedding size of 128 and a hidden size of 256. 
    $D_M$ and $D_S$ are both 1-layer fully connected layers. 
    Since the output of the Encoder is a sequence of $n$ continuous representations, where $n$ is the input sequence length, we do a max pooling operation along the temporal dimension so that we have a fixed vector representation. 
    These fixed vector representations are the direct inputs to $D_M$ and $D_S$.
    The max sequence length is set at 300, and the max phrase length is set at 70. 
    If an input sequence has more than 300 frames, we randomly sample 300 frames at each epoch. 
    If a video in the testing or validation set has more than 300 frames, we fix the indices of the sampled frames to remove any randomness for evaluation. 
    For input sequences that are shorter than 300 frames, we zero-pad the remaining sequence.

\subsection{Loss functions} \label{appendix:loss}

    \subsubsection{Style Disentanglement}

    $D_S$ is trained using Equation \ref{eq:ds loss}. 
    $E_C$ is trained using the same equation, but the labels are flipped and $D_S$ is not updated.
    \begin{equation}\label{eq:ds loss} 
    \mathcal{L}_{{Adv}_{D_S}} = 
    -\mathop{\mathbb{E}} [\log ( D_S(E_C(X_i^s)) )  -  \log (1 - D_S(E_C(X_i^t)) )]
     \end{equation}
     
     \subsubsection{Content Disentanglement}
        $D_C$ is trained by minimizing Equation \ref{eq:dc loss}. 
        $E_S$ is trained by maximizing Equation \ref{eq:es adv loss} with the weights of $D_C$ kept frozen. 
        \begin{equation}\label{eq:dc loss} 
            \mathcal{L}_{{Adv}_{D_C}}  =  - \log p(Y^z_i \vert D_C( E_S(X^f_i) ) ) 
        \end{equation}
        \begin{equation}
        \mathcal{L}_{{Adv}_{E_S}} = H(Y_i^z \vert D_C(E_S(X_i^s)) ) 
        \label{eq:es adv loss} 
        \end{equation}
        
    \subsubsection{Feature Aggregation}
    A Feature Aggregation Module, $M$, combines the disentangled representations from the previous two steps. 
    For any given pair of style and content representations we have:
    \begin{equation}\label{eq:M example} 
        M(z_{style}^f , z_{content}^{f\prime}) = \mathbf{m}( z_{style}^f + z_{content}^{f\prime}) 
    \end{equation}
    
    In Equation \ref{eq:M example}, $\mathbf{m}$ is the LayerNorm operation \cite{ba2016layer}, $f \in \{s,t\}$ and $f^{\prime} \in \{s,t\}$.
    
    \subsubsection{Prediction}
    The Decoder, $G$,  takes in the output of $M$, $h_{f\prime}^f$, and outputs the predicted sentence $\hat{Y}^{f\prime}$, and is trained with the cross-entropy loss using the labels $Y^{f\prime}$. The objective is:
    \begin{equation}\label{cls}
        \mathcal{L}_{cls} = - \log p(Y^{f\prime} \vert G( M(E_S(X^f), E_C(X^{f\prime})) ))
    \end{equation}
    At test time, the model outputs the most likely sentence given a real-life video: 
    \begin{equation}\label{decoding}
         \argmax_{Y} p(Y^{t} \vert G( h^t_t ))
    \end{equation}
    
    \subsubsection{Semantic Alignment}
   
    We create four pairs $\mathcal{G}_k, k \in \{1,2,3,4\} $.
    $\mathcal{G}_1$ and $\mathcal{G}_2$ are outputs of $M$ that share synthetic content: (Synthetic Style, Synthetic Content) and (Real Style, Synthetic Content). 
    $\mathcal{G}_3$ and $\mathcal{G}_4$ share real content: (Synthetic Style, Real Content) and (Real Style, Real Content). 
    A multi-class discriminator, $D_M$, is trained using Equation \ref{DCD_discrim} to correctly identify which group every output of $M$ belongs to. 
    $l_k$ is the corresponding label for a given $\mathcal{G}_k$.
    $E_C$, $E_S$, and $M$ are updated with Equation \ref{DCD_adv} such that $D_M$ can't distinguish outputs of $M$ that are in $\mathcal{G}_1$ and $\mathcal{G}_2$ and outputs of $M$ that are in $\mathcal{G}_3$ and $\mathcal{G}_4$. 
    \begin{equation}\label{DCD_discrim}
        \mathcal{L}_{{Adv}_{D_M}} =
    - \mathop{\mathbb{E}} [\sum\limits_{k=1}^4  l_k \log (D_M(M (\mathcal{G}_k) ))]
    \end{equation}
    
    \begin{equation} \label{DCD_adv}
        \mathcal{L}_{{Adv}_{M}} =
     - \mathop{\mathbb{E}}[ l_1 \log (D_M(M (\mathcal{G}_2) )) \\
     - l_3 \log (D_M(M (\mathcal{G}_4) )) ]
    \end{equation}

    The final loss function to train our model is show in Equation \ref{eq:full loss} where the weightings for each term are tuned using a validation set. 
    An overview of the training procedure is shown in Algorithm \ref{appendix:algo_train}
    \begin{equation}\label{eq:full loss} 
        \mathcal{L} = 
    \lambda_1\mathcal{L}_{cls} + 
    \lambda_2\mathcal{L}_{{Adv}_M} + 
    \lambda_3\mathcal{L}_{{Adv}_{D_M}} + \\
    \lambda_4\mathcal{L}_{{Adv}_{E_S}} + 
    \lambda_5\mathcal{L}_{{Adv}_{D_C}} +
    \lambda_6\mathcal{L}_{{Adv}_{D_S}} 
    \end{equation}
    
    \subsubsection{Hyperparameters}
    
     We use the Adam optimizer with learning rate of 0.0001 for all of our networks. 
    We train for 60k iterations with a batch size of 8, and use a dropout value of 0.15. 
    We use the validation set to tune the different weightings for our loss function. 
    We set $\lambda_1 = 1.0, \lambda_2 = 0.25, \lambda_3 = 1.0, \lambda_4 = 1.0, \lambda_5 = 1.0$ and $\lambda_6 = 1.0$. 
    
\end{document}